%% file: main.tex
\documentclass[letterpaper]{article} % DO NOT CHANGE THIS
\usepackage{aaai23}
\usepackage{times}  % DO NOT CHANGE THIS
\usepackage{helvet}  % DO NOT CHANGE THIS
\usepackage{courier}  % DO NOT CHANGE THIS
\usepackage[hyphens]{url}  % DO NOT CHANGE THIS
\usepackage{graphicx} % DO NOT CHANGE THIS
\usepackage{natbib}  % DO NOT CHANGE THIS AND DO NOT ADD ANY OPTIONS TO IT
\usepackage{caption} % DO NOT CHANGE THIS AND DO NOT ADD ANY OPTIONS TO IT
\DeclareCaptionStyle{ruled}{labelfont=normalfont,labelsep=colon,strut=off} % DO NOT CHANGE THIS
\usepackage{algorithm}
\usepackage{algorithmic}
\usepackage{multirow}
\usepackage{lipsum}
\usepackage{booktabs}
\usepackage{siunitx}
\usepackage{amsmath}
\usepackage{amsfonts}
\usepackage[colorinlistoftodos]{todonotes}

\usepackage{newfloat}
\usepackage{listings}
\DeclareCaptionStyle{ruled}{labelfont=normalfont,labelsep=colon,strut=off} % DO NOT CHANGE THIS
\floatstyle{ruled}
\newfloat{listing}{tb}{lst}{}
\floatname{listing}{Listing}
\title{Financial Time Series Forecasting using CNN and Transformer}

\author{Zhen Zeng, Rachneet Kaur, Suchetha Siddagangappa, \\ Saba Rahimi, Tucker Balch, Manuela Veloso}
\affiliations{%
  J.~P.~Morgan AI Research, New York, NY, USA}
\usepackage{bibentry}
\begin{document}

\maketitle

\begin{abstract}
Time series forecasting is important across various domains for decision-making. In particular, financial time series such as stock prices can be hard to predict as it is difficult to model short-term and long-term temporal dependencies between data points. Convolutional Neural Networks (CNN) are good at capturing local patterns for modeling short-term dependencies. However, CNNs cannot learn long-term dependencies due to the limited receptive field. Transformers on the other hand are capable of learning global context and long-term dependencies.  In this paper, we propose to harness the power of CNNs and Transformers to model both short-term and long-term dependencies within a time series, and forecast if the price would go up, down or remain the same (flat) in the future. In our experiments, we demonstrated the success of the proposed method in comparison to commonly adopted statistical and deep learning methods on forecasting intraday stock price change of S\&P 500 constituents.
\end{abstract}

\input{introduction}

\input{related_work}

\input{data}
\input{method}

\input{experiments}
\input{conclusion}

\bibliography{egbib}

\end{document}

%% file: introduction.tex
\section{Introduction}\label{sec:introduction}

% the problem: time series forecasting
% the literature: where are we
Time series forecasting is challenging, especially in the financial industry~\cite{pedersen2019efficiently}.  It involves statistically understanding complex linear and non-linear interactions within historical data to predict the future. In the financial industry, common applications for forecasting include predicting buy/sell or positive/negative price changes for company stocks traded on the market. 
Traditional statistical approaches commonly adapt linear regression, exponential smoothing~\cite{holt2004forecasting, winters1960forecasting, gardner1985forecasting} and autoregression models~\cite{makridakis2020m4}. With the advances in deep learning, recent works are heavily invested in ensemble models and sequence-to-sequence modeling such as Recurrent Neural Networks (RNNs), Long Short-Term Memory (LSTM)~\cite{hochreiter1997long}.  In Computer Vision domain, Convolutional Neural Networks (CNN)~\cite{ren2015faster, ronneberger2015u} have shown prominence in learning local patterns which are suitable for modeling short-term dependencies, although not suitable for modeling long-term dependencies due to limited receptive field.  Most recently, Transformers~\cite{vaswani2017attention}, have shown great success in Natural Language Processing (NLP)~\cite{devlin2018bert, brown2020language, smith2022using} domain, achieving superior performance on long-term dependencies modeling compared to LSTM.

% the contribution: two folds
Our contributions is the following: we leverage the advantages of CNNs and Transformers to model short-term and long-term dependencies in financial time series, as shown in Figure~\ref{fig:overview}. In our experiments, we show the advantage of the proposed approach on intraday stock price prediction of S\&P 500 constituents, outperforming statistical methods including Autoregressive Integrated Moving Average (ARIMA) and Exponential Moving Average (EMA) and a state-of-the-art deep learning-based autoregressive model DeepAR~\cite{salinas2020deepar}.

%% file: related_work.tex
\section{Related Works}
\begin{figure*}[h!]
    \centering
    \includegraphics[scale = 0.31]{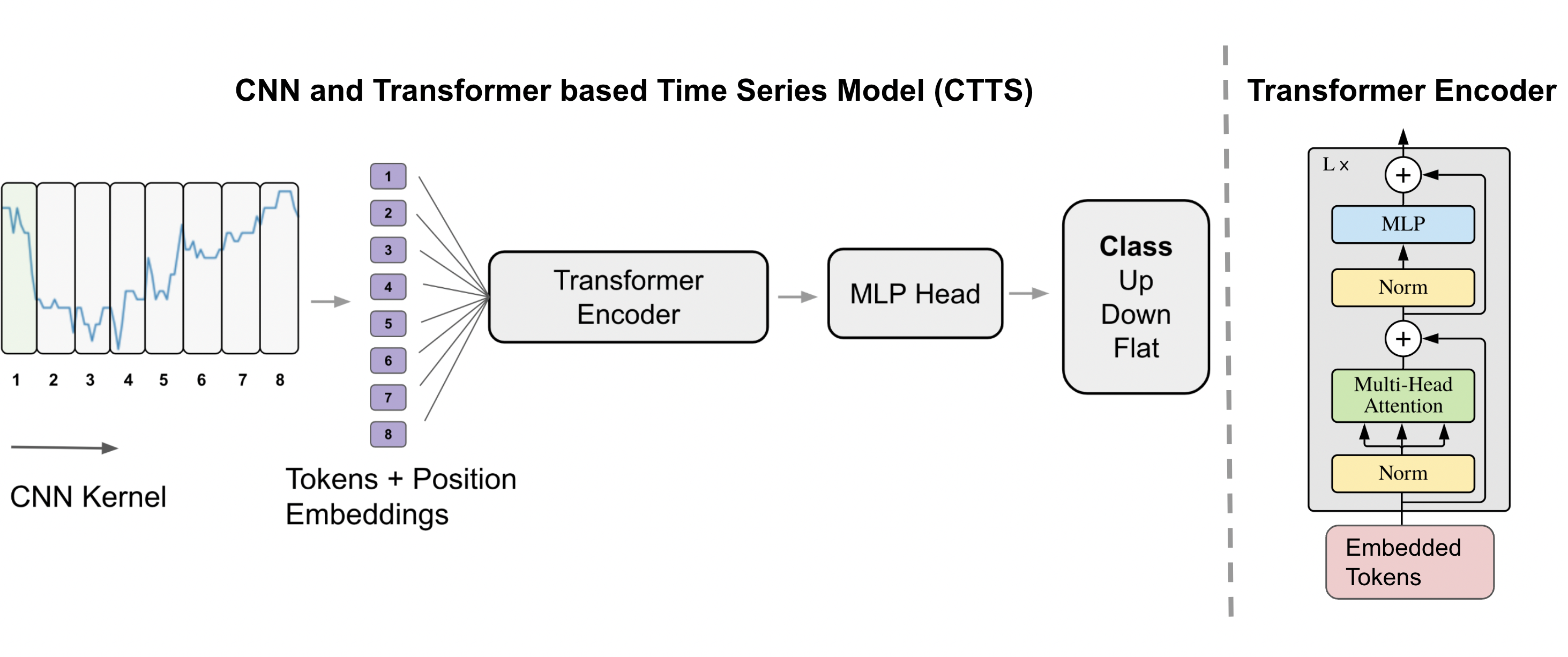}
    \caption{Overview of the proposed approach. Quick peak of the transformer encoder architecture on the right~\cite{dosovitskiy2020image}}
    \label{fig:overview}
\end{figure*}

\subsection{Time series forecasting}\label{ts_forecasting} Typical forecasting techniques in the literature utilize statistical tools, such as, exponential smoothing (ETS) ~\cite{holt2004forecasting, winters1960forecasting, gardner1985forecasting} and autoregressive integrated moving average (ARIMA) ~\cite{makridakis2020m4}, on numerical time series data for making one-step-ahead predictions. These predictions are then recursively fed into the future inputs to obtain multi-step forecasts. Multi-horizon forecasting methods such as ~\cite{taieb2010multiple, marcellino2006comparison} directly generate simultaneous predictions for multiple pre-defined future time steps.

\subsubsection{Machine learning and deep learning based approaches}\label{ml_dl_ts_forecasting}
%Why is ML better than statistical approaches 
Machine learning (ML) approaches have shown to improve performance by addressing high-dimensional and non-linear feature interactions in a model-free way. These methods include tree-based algorithms, ensemble methods, neural network,  autoregression and recurrent neural networks \cite{hastie2001elements}. More recent works have applied Deep learning (DL) methods on numeric time series data \cite{bao2017deep, gensler2016deep, romeu2015stacked, sagheer2019time, Sutskever2014sequence}.
%Why is DL better than ML approaches 
DL automates the process of feature extraction and eliminates the need for domain expertise.

Since the introduction of transformers \cite{vaswani2017attention}, they have become the state of the art model to improve the performance of NLP applications. The commonly used approach is to pre-train on a large dataset and then fine-tune on a smaller task-specific dataset \cite{devlin2018bert}. Transformers leverage from multi-headed self-attention and replace the recurrent layers most commonly used in encoder-decoder architectures. In contrast to RNNs and LSTMs where the input data is sequentially processed, transformers bypass the recursion to ingest all inputs at once; thus, transformers allow for parallel computations to reduce training time and do not suffer from long-term memory dependency issues.
%Add LSTM seq to seq 

%Briefly describe our work 

%Structure for the rest of the paper 
The remainder of the paper is organized as follows. We discuss our proposed methodology for time series modeling  
% (\ref{subsec:fine-tune}). 
% In section \ref{sec:experiments}, 
Further, we discuss our benchmark baseline models along with the performance evaluation metrics and report the experimental results. Finally, 
% in section \ref{sec:conclusion}, 
we highlight some concluding remarks and future directions for this study.

%Related works - Prepare code to visualize attention using bars on original price time series 

%% file: data.tex
% \section{Data}\label{sec:data}
% \begin{figure*}[htb]
%     \centering
%     \includegraphics[scale = 0.45]{figures/data_visual_new.png}
%     \caption{Examples for stock intraday price time series (top), processed return as in Equation~\ref{eq:return} (middle) and time frequency spectrogram  (bottom) from S\&P 500. Note that each time step here denotes a 6-second time interval.}% The image patch row at the top of each spectrogram is the normalized return time series represented through intensity of the pixels, i.e., mapped from floats in range [-1, 1] to integers in range [0, 255]. Note that each time step here denotes a 6-second time interval. }
%     \label{fig:data_examples}
% \end{figure*}

% We used the intraday stock prices of S\&P 500 constituent stocks obtained from licensed Bloomberg data service~\cite{bloomberg}. The data was sampled at 1 minute interval for the year 2019 (52 weeks, each week has at most 5 trading days). For every stock, we sampled 7 time series for each stock each week.

% \todo[inline]{Change the statistics} Every time series contains 85 time steps.  Overall, we had around 833k time series.

% @book{pedersen2019efficiently,
%   title={Efficiently inefficient: how smart money invests and market prices are determined},
%   author={Pedersen, Lasse Heje},
%   year={2019},
%   publisher={Princeton University Press}
% }

%% file: method.tex
\section{Method}\label{sec:method}

Our proposed method is called
\underline{\textbf{C}}NN and \underline{\textbf{T}}ransformer based \underline{\textbf{t}}ime \underline{\textbf{s}}eries modeling (CTTS) as shown in overview Figure~\ref{fig:overview}.

\subsection{Preprocessing}\label{subsec:preprocess}
We used standard min-max scaling to standardize each input time series within $[0, 1]$. Given a raw stock prices time series $\textbf{x}$, the standardized time series were calculated as 
\begin{equation*}
    \textbf{x}_{standardized} = \frac{\textbf{x} - \min(\textbf{x})}{\max(\textbf{x}) - \min(\textbf{x})}
\end{equation*}
This $\textbf{x}_{standardized}$ is then passed into our model to learn the sign of the change in the very next time step, which is defined as our prediction target. 

\subsection{Forecasting}\label{subsec:fine-tune}
As shown in Figure~\ref{fig:overview}, we use 1D CNN kernels to convolute through a time series, projecting each local window into an embedding vector that we call a token. Each token carries the short-term patterns of the time series.  we then add positional embedding~\cite{vaswani2017attention} to the token and pass that through a Transformer model to learn the long-term dependencies between these tokens. The transformer model outputs a latent embedding vector of the time series, which is then passed through a Multilayer Perceptron (MLP) with softmax activation in the end to generate sign classification outputs. The output is in the form of probability for each class (up, down, flat), where the probability for all 3 classes sum up to 1.

%% file: experiments.tex
\section{Experiments}\label{sec:experiments}
In our experiments, we benchmarked  CTTS - our proposed method against 4 methods.  1) DeepAR - a state-of-the-art autoregressive recurrent neural networks-based time series forecasting method, 2) AutoRegressive Integrated Moving Average (ARIMA), 3) Exponential Moving Average (EMA) and 4) naive constant class predictions.  
We benchmarked the performance using sign prediction accuracy and thresholded version of the sign prediction accuracy (discussed later) using our model prediction probabilities. 
We ran our experiments on a Linux machine with 8 16GB NVIDIA T4 Tensor Core GPUs, and using PyTorch v1.0.0 DL platform in Python 3.6. In all models, we set a fixed random seed for reproducible results.

\begin{table}[t!] 
\centering
\begin{tabular}{|>{\centering\arraybackslash} m{1.1cm}| >{\centering\arraybackslash} m{1.1cm} | >{\centering\arraybackslash} m{1.3cm} | >{\centering\arraybackslash} m{1.1cm} | >{\centering\arraybackslash} m{1.3cm} |} 
\hline
 Method &
\begin{tabular}[c]{@{}c@{}}2-class $\uparrow$ \end{tabular} &
\begin{tabular}[c]{@{}c@{}}2-class$^*$ $\uparrow$\end{tabular}
&
\begin{tabular}[c]{@{}c@{}}3-class $\uparrow$ \end{tabular}
&
\begin{tabular}[c]{@{}c@{}}3-class$^*$  $\uparrow$ \end{tabular}
\\ \hline\hline
EMA & 53.2\% & 59.9\% & 39.5\% & 41.7\% \\ 
\hline
ARIMA & 50.9\% & 51.8\% & 37.5\% & 38.4\% \\ 
\hline
DeepAR & 51.1\% & 53.6\% & 37.4\% & 38.7\% \\ 
\hline
\textbf{CTTS} & \textbf{56.7\%} & \textbf{66.8\%} & \textbf{44.1\%} & \textbf{55.2\%} \\ 
\hline
\end{tabular}
\vspace{1mm}
\caption{Summary of sign accuracy over the last quarter of 2019. Our proposed method CTTS outperforms the baselines DeepAR, ARIMA and EMA. 2/3-class$^*$ refers to the thresholded version of the sign accuracy. The distribution of signs in the ground truth test set are as follows: price goes up $|$ down $|$ remains flat: 37.1\% $|$ 36.5\% $|$ 26.4\%, respectively. If naively predicting the majority class (up) for all time series, the sign accuracy will only be 37.1\%.}
\label{tab:sign_accuracy_results}
\end{table}

\subsection{Experimental setup}\label{subsec:setup}
\subsubsection{Data}
We used the intraday stock prices of S\&P 500 constituent stocks obtained from licensed Bloomberg data service~\cite{bloomberg}. The data was sampled at 1 minute interval for the year 2019 (52 weeks, each week has 5 trading days). For every stock, we sampled 7 time series for each week. Data from the first three quarters (weeks 1 to 39) were used for training and validation.  Data was randomly split and 80\% was used for training and the remaining 20\% for validation. We had around 507K training and 117K validation samples. Data from the last quarter (weeks 40 to 52) was used for testing, totaling 209K time series. For each time series, the first 80 time steps (\emph{input}) were used to forecast the sign of price change at the 81st time step (\emph{target}). The overall test set performance is denoted by the aggregate for the evaluation metrics over the test set. 

\subsubsection{CTTS}\label{subsubsec:ctts}
In our experiments, we used cross entropy loss, and AdamW~\cite{loshchilov2017decoupled} optimizer with batch size of 64 and max epoch of 100. we used CNN kernel size of 16 and stride 8. Transformer with depth 4 and 4 self-attention heads, with an embedding dimension of 128, and a drop rate of 0.3 to prevent overfitting. 

\subsection{Baselines}
% \zz{details on architecture, and training details on loss, learning rate, epochs, optimizer, weight decay, batch size.}
% \input{vits}
% \input{nts}
\input{baselines}

\subsection{Metric}
We evaluated all our models on 3-class [class 1-price goes up, class 2-price goes down, class 3-price remains flat] and 2-class [class 1-price goes up or remains flat, class 2-price goes down]. We used the averaged sign prediction accuracy over all test samples to evaluate our models. Higher accuracy implies more accurate forecasts. 

%TO DO: Threshold sign accuracy 
Further, we evaluate a thresholded version of the sign accuracy, where we defined our threshold as the 75$^{th}$ percentile of all predicted probabilities of dominating classes. We retain only the samples that exceed the threshold and compute the sign prediction accuracy over these retained samples. 

\subsection{Results \& Discussion}
We summarize the quantitative benchmark results from the sign accuracy in Table~\ref{tab:sign_accuracy_results}. We show the sign accuracy for both 2-class and 3-class tasks as explained above. Note that random guess in 2-class task leads to 50\% of accuracy, and random guess in 3-class task leads to 33\% of accuracy. As shown in the 2-class and 3-class columns of Table~\ref{tab:sign_accuracy_results}, CTTS outperformed all baseline methods in both cases. This demonstrates the benefit of combining CNNs and transformers for time series forecasting. 

Further, we evaluate a thresholded version of the sign accuracy, where we defined our threshold as the 75th percentile of all predicted probabilities of dominating classes. We retain only the samples that exceed the threshold and compute the sign prediction accuracy over these retained samples. As shown in 2-class$^{*}$ and 3-class$^*$ columns of Table~\ref{tab:sign_accuracy_results}, the accuracy after thresholding over the prediction probability has increased for all methods. In addition, the gain of accuracy increase is the most for our proposed method CTTS. This shows that the class probabilities that CTTS outputs are reliable. Specifically, high-confidence predictions from CTTS are often correct, thus thresholding filters out erroneous low-confidence predictions, leading to a boost in the sign accuracy.

% showing that CTTS is calibrated in that the predicted class probability represents well the confidence level of the prediction, such that high confidence predictions often correspond to correct predictions 

Another highlight is the significant gap that CTTS has achieved for thresholded 3-class$^*$ accuracy compared to other baselines. This can be harnessed in the future for trading decision-making. For example, a straightforward trading decision can be buy/sell/hold stocks when the predicted class is up/down/flat, respectively. Given that CTTS's predicted probablitlies are reliable as discussed earlier, the amount of stock shares to buy/sell/hold can depend on the predicted probability, the higher the probability, the more shares to consider. 

% Further, we demonstrate the profit factor from our discussed trading strategy utilizing model predictions in Table \ref{tab:profit_factor_results}. Note that the profit factor from the buy all and hold all trading strategy is the same; this is because both gross profit and loss (in \$) in the buy all strategy is double the gross profit and loss in the hold all case, keeping the profit factor unchanged. 

% The profit factor is computed per stock ticker-week pair, 
% For each week, for each ticker
% aggregate d results are mean across tickes, average profit factor across SNP 500 and std is across the weeks. 

%% file: baselines.tex
\subsubsection{DeepAR}\label{subsubsec:baselines}
%Add how do we define classes or signs from continuous predictions of DeepAR and how do we aggregate over 200 samples  
DeepAR forecasting algorithm is a supervised learning algorithm for forecasting 1D time series using autoregressive recurrent neural networks. DeepAR benefits from training a single model jointly over all of the time series. We used a batch size of 128, Adam optimizer, and the normal distribution loss. The model was trained for a maximum of 300 epochs, with early stopping mechanism set to a patience of 15. The base learning rate was 1e-3, adjusted by learning rate scheduler with decay factor 0.1 and a patience of 5. We also used dropout with probability 0.1. DeepAR generated multiple (200) samples of the prediction target for each time series and we defined the prediction probability over the three classes as the proportion of samples predicted per class.

\subsubsection{ARIMA}
Autoregressive Integrated Moving Average (ARIMA)
models capture autocorrelations in the data using a combination approach of autoregressive model, moving average model, and differencing \cite{wilks2011statistical}. We compared the continuous valued ARIMA forecasts with the last known price in the input data to define the predicted sign, and the corresponding prediction probability was defined as the percentage of the absolute delta between the predicted and the last known price with respect to the standard deviation of the past 80 data points, capped by 1.

\subsubsection{EMA} 
An exponential moving average (EMA) is a type of moving average that places a greater weight and significance on the most recent data points. We used the "estimated" initialization method, which treats the initial values like parameters, and chooses them to minimize the sum of squared errors. Similar to ARIMA, we delta between the predicted and the last known price to define the predicted sign, and the corresponding prediction probability.

\subsubsection{Constant Class Prediction}
We tested the three naive baselines where we always predict a constant sign i.e., either the predicted price always goes up, down, or remains flat. The prediction probabilities for these were set to 1 for the winning class, and 0 for the remaining two classes.

%% file: conclusion.tex
\section{Conclusion}\label{sec:conclusion}
In this paper, we tackle the challenging problem of time series forecasting of stock prices in the financial domain. In this paper, we demonstrated the combined power of CNN and Transformer to model both short-term and long-term dependencies within a time series. In our experiments over intraday stock price of S\&P 500 constituents in year 2019, we demonstrated the success of the proposed method CTTS in comparison to ARIMA, EMA, and DeepAR, as well as the potential for using this method for downstream trading decisions in the future.